\documentclass[letterpaper, 10pt, conference]{ieeeconf}

\IEEEoverridecommandlockouts                              

\usepackage{pgfplots}
\pgfplotsset{compat=newest}
\usetikzlibrary{plotmarks}
\usetikzlibrary{arrows.meta}
\usepgfplotslibrary{patchplots}
\usepackage{multicol}
\usepackage[fleqn]{amsmath}
\usepackage{booktabs}
\usepackage{amssymb}
\usepackage{color}
\usepackage{graphicx}
\usepackage{pstricks}
\usepackage{comment}
\usepackage{mathtools}
\usepackage{setspace}
\usepackage{breqn}
\usepackage{makecell}
\usepackage{multirow}
\usepackage{afterpage}
\usepackage{subcaption}
\usepackage{siunitx}
\usepackage{import}
\let\labelindent\relax
\usepackage{enumitem}
\usepackage[noadjust]{cite}
\makeatletter
\let\NAT@parse\undefined
\makeatother
\usepackage[hidelinks,bookmarks=true]{hyperref}

\allowdisplaybreaks

\DeclareMathOperator{\Beta}{Beta}
\DeclareMathOperator{\RMSE}{RMSE}

\newcommand{\transpose}[1]{#1^\intercal}

\newcommand{\norm}[1]{\left\lVert #1 \right\rVert}
\newcommand{\mean}[1]{\mathrm{E}[#1]}
\newcommand{\var}[1]{\mathrm{var}[#1]}

\def\atan2{\mathop{\rm atan2}}

\title{\LARGE \bf Long-Term Urban Vehicle Localization Using Pole Landmarks Extracted from 3\nobreakdash-D Lidar Scans}

\author{
	Alexander~Schaefer, Daniel~B{\"u}scher, Johan~Vertens, Lukas~Luft, Wolfram~Burgard
	\thanks{\copyright\ 2019 IEEE. Personal use of this material is permitted.  Permission from IEEE must be obtained for all other uses, in any current or future media, including reprinting/republishing this material for advertising or promotional purposes, creating new collective works, for resale or redistribution to servers or lists, or reuse of any copyrighted component of this work in other works.}
	\thanks{This work has been partially supported by Samsung Electronics~Co.~Ltd. under the GRO~program.}%
	\thanks{All authors are with the Department of Computer Science, University of Freiburg, Germany.}%
	\thanks{\tt \footnotesize \{aschaef, buescher, vertensj, luft, burgard\} @cs.uni-freiburg.de}
	\thanks{978-1-7281-3605-9/19/\$31.00 \textcopyright 2019 IEEE}%
}%

\begin{document}

\maketitle

\begin{abstract}
Due to their ubiquity and long-term stability, pole-like objects are well suited to serve as landmarks for vehicle localization in urban environments.
In this work, we present a complete mapping and long-term localization system  based on pole landmarks extracted from 3\nobreakdash-D lidar data.
Our approach features a novel pole detector, a mapping module, and an online localization module, each of which are described in detail, and for which we provide an open-source implementation~\cite{polex2019}.
In extensive experiments, we demonstrate that our method improves on the state of the art with respect to long-term reliability and accuracy:
First, we prove reliability by tasking the system with localizing a mobile robot over the course of 15~months in an urban area based on an initial map, confronting it with constantly varying routes, differing weather conditions, seasonal changes, and construction sites.
Second, we show that the proposed approach clearly outperforms a recently published method in terms of accuracy.
\end{abstract}

\section{Introduction}
\label{sec:introduction}

Intelligent vehicles require accurate and reliable self-localization systems.
Accurate, because an exact pose estimate enables complex functionalities such as automatic lane following or collision avoidance in the first place.
Reliable, because the quality of the pose estimate must be maintained independently of environmental factors in order to ensure safety.

Satellite-based localization systems like RTK-GPS or DGPS seem to be an efficient solution, since they achieve centimeter-level accuracy out of the box.
However, they lack reliability.
Especially in urban areas, buildings that obstruct the line of sight between the vehicle and the satellites can decrease accuracy to several meters~\cite{modsching2006, carlevaris2015}.
Localization on the basis of dense maps like grid maps, point clouds, or polygon meshes represents a more reliable alternative~\cite{levinson2010}.
On the downside, dense approaches require massive amounts of memory that quickly become prohibitive for maps on larger scales.
This is where landmark maps come into play: 
By condensing billions of raw sensor data points into a comparably small number of salient features, they can decrease the memory footprint by several orders of magnitude~\cite{kuemmerle2019}.

In this work, we present an approach to long-term 2\nobreakdash-D vehicle localization in urban environments that relies on pole landmarks extracted from mobile lidar data.
Poles occur as parts of street lamps, traffic signs, as bollards and tree trunks.
They are ubiquitous in urban areas, long-term stable and invariant under seasonal and weather changes.
Since their geometric shape is well-defined, too, poles are well suited to serve as landmarks that enable accurate and reliable localization.

\begin{figure}
	\centering
	\resizebox{\columnwidth}{!}{\input{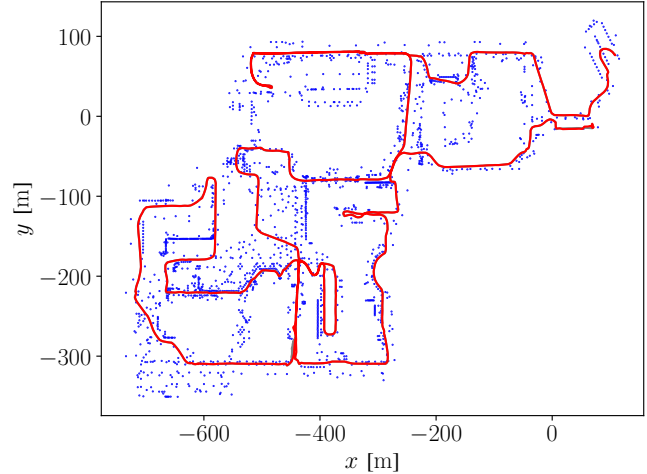}}
	\caption{
		Pole landmark map created from the NCLT dataset~\cite{carlevaris2015} and trajectory of an experimental run 15~months after map creation.
		The blue dots represent the landmarks.
		The gray line corresponds to the ground-truth trajectory.
		Most of it is covered by the red line, which represents the estimate produced by the presented method.
		The mean position difference between both trajectories, formally defined in section~\ref{subsec:localization_nclt_dataset}, amounts to \SI{0.31}{m}.}
	\label{fig:nclt_localization}
\end{figure}

Our localization process is subdivided into a mapping and a localization phase.
During mapping, we use the pole detector presented below to extract pole landmarks from lidar scans and register them with a global map via a given ground-truth vehicle trajectory.
During localization, we employ a particle filter to estimate the vehicle pose by aligning the pole detections from live sensor data with those in the map.
Figure~\ref{fig:nclt_localization} shows an exemplary localization result.

We are not the first ones to propose this kind of localization technique:
The next section provides an overview over the numerous related works.
However, to the best of our knowledge, we are the first ones to present a pole detector that does not only consider the laser ray endpoints, but also the free space in between the laser sensor and the endpoints, and to demonstrate reliable and accurate vehicle localization based on a map of pole landmarks on large time scales.
While related works usually evaluate localization performance on a short sample trajectory of at most a few minutes length, we successfully put our approach to the test on a publicly available long-term dataset that contains 35~hours of data recorded over the course of 15~months -- including varying routes, construction zones, seasonal and weather changes, and lots of dynamic objects.
Additional control experiments show that the presented method is not only reliable, but significantly outperforms a recently published state-of-the-art approach in terms of accuracy, too.

\section{Related Work}
\label{sec:related_work}

In recent years, a number of authors have addressed the specific question of vehicle localization via pole landmarks extracted from lidar scans.
Any solution to this question consists of at least two parts: a pole detector and a landmark-based pose estimator.
The detector developed by Weng et~al.~\cite{weng2018}, for example, tessellates the space around the lidar sensor and counts the number of laser reflections per voxel.
Poles are then assumed to be located inside contiguous vertical stacks of voxels that all exceed a reflection count threshold.
In order to extract the pole parameters from these clusters, the detector fits a cylinder to all points in a stack via RANSAC~\cite{fischler1981}.
For 2\nobreakdash-D pose estimation, the authors employ on a particle filter with nearest-neighbor data association.
Sefati et~al.~\cite{sefati2017} present a pole detector that removes the ground plane from a given point cloud, projects the remaining points onto a horizontal regular grid, clusters neighboring cells based on occupancy and height, and fits a cylinder to each cluster.
Like Weng et~al., Sefati et~al. obtain their 2\nobreakdash-D localization estimate from a particle filter that performs nearest-neighbor data association.
K{\"u}mmerle et~al.~\cite{kuemmerle2019} make use of Sefati et~al.'s pole detector, but to further refine the localization estimate, they also fit planes to building fa\c{c}ades in the laser scans and lines to lane markings in stereo camera images.
Like the above works, their pose estimator relies on a Monte Carlo method to solve the data association problem, but uses optimization to compute the most likely pose.
More specifically, in the data association stage, it builds a local map by accumulating the landmarks detected over the past timesteps based on odometry.
It then samples multiple poses around the current GPS position, uses these pose hypotheses to project the local map into the global map, and identifies the most probable hypothesis via a handcrafted landmark matching metric.
Given the resulting data associations, it refines the current vehicle pose estimate via nonlinear least squares optimization over a graph of past vehicle poses and landmarks.

Spangenberg et~al.~\cite{spangenberg2016} extract pole landmarks not from lidar scans, but from stereo camera images.
In order to estimate the vehicle pose, they feed wheel odometry, GPS data, and online pole detections to a particle filter.

While the approaches above all provide a complete localization system consisting of a pole extractor and a landmark-based localization module, there exist a variety of research papers that focus solely on pole extraction.
Extracting poles from lidar data is a common problem in road infrastructure maintenance and urban planning.
In this domain, researchers are not only interested in fitting geometric primitives to the data and determining pole coordinates, but also in precise point-wise segmentation.
Brenner~\cite{brenner2009}, Cabo et~al.~\cite{cabo2014}, Tombari et~al.~\cite{tombari2014}, and Rodriguez et~al.~\cite{rodriguez2015} present different methods to extract pole-like objects from point clouds, i.e. without accounting for free space information.
The approaches of Yu et~al.~\cite{yu2015} and Wu et~al.~\cite{wu2017} specifically target street lamp poles, while Zheng et~al.~\cite{zheng2016} provide a solution to detect poles that are partially covered by vegetation.
Yokoyama et~al.~\cite{yokoyama2013} not only extract poles, but they classify them as lamp posts, utility poles, and street signs.
Ord{\'o}{\~n}ez et~al.~\cite{ordonez2017} build upon the pole detector proposed by Cabo et~al.~\cite{cabo2014} and classify the results into six categories, including trees, lamp posts, traffic signs, and traffic lights.
Li et~al.~\cite{li2018} take classification one step further by decomposing multifunctional structures, for example a light post carrying traffic signs, into individual elements.

Poles are not the only landmarks suitable for vehicle localization.
Qin et~al.~\cite{qin2012} investigate Monte Carlo vehicle localization in urban environments based on curb and intersection features.
As demonstrated by the works of Schindler~\cite{schindler2013} and Schreiber et~al.~\cite{schreiber2013}, road markings as landmarks can also yield high localization accuracy.
Hata and Wolf~\cite{hata2016} feed both curb features and road markings to their particle filter.
Welzel et~al.~\cite{welzel2015} explore the idea of using traffic signs as landmarks.
Although traffic signs occur less frequently in urban scenarios compared to other types of road furniture like road markings or street lamp poles, they offer the advantage of not only encoding a position, but also an unambiguous ID.
Finally, Im et~al.~\cite{im2016} explore urban localization based on vertical corner features, which appear at the corners of buildings, in monocular camera images and lidar scans.

\section{Approach}
\label{sec:approach}

The proposed 2\nobreakdash-D vehicle localization system consists of three modules: the pole extractor, the mapping module, and the localization module.
During the initial mapping phase, the pole extractor reduces a given set of lidar scans to pole landmarks.
The mapping module then uses the ground-truth sensor poses to build a global reference map of these landmarks.
During the subsequent localization phase, the pole extractor processes live lidar data and passes the resulting landmarks to the localization module, which in turn generates a pose estimate relative to the global map.
In the following, we detail each of these modules and their interactions.

\subsection{Pole Extraction}
\label{subsec:pole_extraction}

The pole extraction module takes a set of registered 3\nobreakdash-D lidar scans as input and outputs the 2\nobreakdash-D coordinates of the centers of the detected poles with respect to the ground plane, along with the estimated pole widths.
To that end, it builds a 3\nobreakdash-D occupancy map of the scanned space, applies a pole feature detector to every voxel, and regresses the resulting pole map to a set of pole position and width estimates.

To describe these three steps mathematically, we denote a single laser measurement -- a ray -- by \mbox{$z \coloneqq \{u, v\}$}, where $u$ and $v$ represent its Cartesian starting point and endpoint, respectively.
All measurements \mbox{$Z \coloneqq \{z_i\}$} are assumed to be registered with respect to the map coordinate frame, whose $x$-$y$ plane is aligned with the ground plane.
The measurements can be taken at different points in time, but the timespan between the first and the last measurement needs to be sufficienctly small in order not to violate our assumption that the world is static.
Now, we tessellate the map space, trace the laser rays, and model the posterior probability that the $j$-th voxel reflects an incident laser ray according to Luft et~al.~\cite{luft2017} by
\begin{align*}
p(\mu_j \mid Z) = \Beta(h_j + \alpha, m_j + \beta).
\end{align*}
Here, $h_j$ and $m_j$ denote the numbers of laser reflections and transmissions in the $j$-th cell, whereas $\alpha$ and $\beta$ are the parameters of the prior reflection probability \mbox{$p(\mu_j) = \Beta(\alpha, \beta)$}, which we determine in accord with \cite{luft2017} by
\begin{align*}
\alpha &= -\frac{\gamma (\gamma^2 - \gamma + \delta)}
{\delta}, &
\beta  &= \frac{\gamma - \delta + \gamma \delta 
	- 2 \gamma^2 + \gamma^3}{\delta},
\end{align*} 
where \mbox{$M \coloneqq \{h_j (h_j + m_j)^{-1}\}$} denotes the maximum-likelihood reflection map, and where \mbox{$\gamma \coloneqq \mean{M}$}, \mbox{$\delta \coloneqq \var{M}$} represent its mean and variance, respectively.
Please note that \mbox{$\{p(\mu_j \mid Z)\}$} is a full posterior map:
In contrast to $M$, which assigns each voxel the most probable reflection rate, it yields a posterior distribution over every reflection rate possible.

Since we want to extract poles based on occupancy probability, not on reflection rate, we convert \mbox{$\{p(\mu_j \mid Z)\}$} to an occupancy map~\mbox{$O \coloneqq \{o_j\}$}.
Assuming that a cell is occupied if its reflection rate exceeds a threshold $\mu_o$, we formulate the occupancy probability by integration: 
\begin{align*}
\mbox{$o_j \coloneqq \int_{\mu_o}^1 p(\mu_j \mid Z)\ d\mu_j$}.
\end{align*}

Next, a pole feature detector transforms $O$ to a 2\nobreakdash-D map of pole scores~$S$ in the ground plane.
Each pixel of $S$ encodes the probability that a pole is present at the corresponding location.
The transformation from $O$ to $S$ follows a set of heuristics that are based on the definition of a pole as a vertical stack of occupied voxels with quadratic footprint, laterally surrounded by a hull of free voxels.
First, we create a set of intermediate 3\nobreakdash-D score maps of the same size as $O$, each denoted by $Q_a \coloneqq \{q_{a,j}\}$.
Every cell~$q_{a, j}$ tells how probable it is that this portion of space is part of a pole with edge length~$a$, where \mbox{$a \in \mathbb{N}^+$} is measured in units of grid spacing:%
{%
\medmuskip=1mu%
\thinmuskip=1mu%
\thickmuskip=1mu%
\begin{align*}
\label{eq:q}
	q_{a, j} \coloneqq  \max_{k \in \mathrm{inside}(j, a)} 
		\Bigg( 
				\frac{\sum \limits_{l \in \mathrm{inside}(k, a)} o_l}{a^2} 
					- \max_{l \in \mathrm{outside}(k, a, f)} o_l 
		\Bigg).
\end{align*}%
}%
Here, $\mathrm{inside}(j, a)$ and $\mathrm{outside}(j, a, f)$ are functions that, given a map index~$j$ and a pole width~$a$, return a set of indices into voxels in the same horizontal map slice as $j$.
While the former outputs the indices of all voxels inside the pole, the latter returns the indices corresponding to the supposedly free region around the pole with thickness~\mbox{$f \in \mathbb{N}^+$}.
Both functions assume that the lower left lateral walls of the pole are aligned with the lower left lateral sides of the $j$-th voxel.
With these definitions, the argument of the enclosing maximum operator amounts to the difference between the mean occupancy value inside the pole and the maximum occupancy value of the volume of free space around the pole.
The resulting score lies in the interval~\mbox{$[-1, 1]$}: the higher the score, the greater the probability that the corresponding partition of space is part of a pole.
Second, we regress from the resulting 3\nobreakdash-D maps~$\{Q_a\}$ to 2\nobreakdash-D by merging them into a single map \mbox{$Q \coloneqq \{q_j\} \coloneqq \{\max_a q_{a, j}\}$} and by determining for each horizontal position in $Q$ the contiguous vertical stack of voxels that all surpass a given score threshold~$q_{\min}$.
After discarding all stacks that fall below a certain height threshold~$h_{\min}$ and computing the mean score for each of the remaining stacks, we obtain the desired 2\nobreakdash-D score map~$S$.

Finally, we convert this discrete score map to a set of continuous pole position and width estimates.
We identify the pole positions as the modes of $S$, which we determine via mean shift~\cite{fukunaga1975} with a Gaussian kernel and with the local maxima of $S$ as seed points.
The width estimate of each pole is computed as the weighted average over all pole widths~$a$, where for every $a$, the weight is the mean of all cells in $Q_a$ that touch the pole.

The presented algorithm differs from other pole extractors in the fact that it is based on ray tracing.
By considering not only the scan endpoints, but also the starting points, it explicitly models occupied and free space.
In contrast, most other methods assume the space around the sensor to be free as long as it does not register any reflections.
The absence of reflections, however, can have two reasons: 
The respective region is in fact free, or the lidar sensor did not cover region due to objects blocking its line of sight or its limited range.

\subsection{Mapping}
\label{subsec:mapping}

In theory, the global reference map could be built by simply applying the pole extractor to a set of registered laser scans that cover the area of interest.
In practice, the high memory complexity of grid maps and laser scans often renders this naive approach infeasible.
To create an arbitrarily large landmark map with limited memory resources, we partition the mapping trajectory into shorter segments of equal length and feed the lidar measurements taken along each segment to the pole extractor one by one.
For the sake of consistency, we take care that the intermediate local grid maps are aligned with the axes of the global map and that all of them have the same raster spacing.
The intermediate maps, whose sizes are constant and depend on the sensor range,  usually fit into memory easily.
Processing all segments provides us with a set of pole landmarks.
If the length of a trajectory segment is smaller than the size of a local map, the local maps overlap, a fact that can lead to multiple landmarks representing a single pole.
In order to merge these ambiguous landmarks, we project all poles onto the ground plane, yielding a set of axis-aligned squares.
If multiple squares overlap, we reduce them to a single pole estimate by computing a weighted average over their center coordinates and widths.
Each weight equals the mean pole score, which we determine by averaging over the scores of all voxels that touch the pole in all score maps $Q_a$.
If there is no overlap, we integrate the corresponding pole into the global reference map without further ado.

As a side benefit, this mapping method allows us to filter out dynamic objects at the landmark level using a sliding-window approach:
A local landmark is integrated into the reference map only if it was seen at least $c$~times in the past $w$~local maps, where \mbox{$c \leq w$}; \mbox{$c, w \in \mathbb{N}^+$}.
Correspondences between landmarks are again determined via checking for overlapping projections in the ground plane.

\subsection{Localization}
\label{subsec:localization}

During online localization, we continuously update the vehicle pose based on the collected odometry measurements and periodically correct the estimate by matching online pole landmarks, which we extract from the most recent local map, against the reference map.
We build the local map by accumulating laser scans along a segment of the trajectory and by registering them via odometry.
To filter out dynamic objects, we apply the sliding-window approach described in the previous section.

A particle filter is well suited for the localization task~\cite{thrun2005}, because it can not only maintain multiple pose hypotheses in parallel, but also handle global localization.
At time~$t$, each particle corresponds to a 2\nobreakdash-D vehicle pose hypothesis, represented by the \mbox{$3 \times 3$} homogeneous transformation matrix~$X_t$.
To perform the motion update, we assume Gaussian motion noise~$\Sigma$ and sample from a trivariate normal distribution in $\chi$:
\begin{align*}
X_{t} = \mathrm{transform}(\xi)\ X_{t-1} \quad
	\big| \quad \xi &\sim \mathcal N(\chi, \Sigma),
\end{align*}
where \mbox{$\chi \coloneqq \transpose{[x, y, \phi]}$} denotes the latest relative odometry measurement, with $x$, $y$, and $\phi$ representing the translation and the heading of the vehicle, respectively.
The function \mbox{$\mathrm{transform}(\transpose{[x, y, \phi]})$} converts the input vector to the corresponding \mbox{$3 \times 3$} transformation matrix.
In each measurement update, we determine the data associations between the online landmarks~\mbox{$\Lambda \coloneqq \{\lambda_k\}$} and the landmarks in the reference map~\mbox{$L \coloneqq \{l_n\}$} via nearest-neighbor search in a $k$\nobreakdash-D tree, assume independence between the elements of $\Lambda$, and update the particle weights according to the measurement probability
\begin{align*}
p(\Lambda \mid X, L) = \prod \limits_{k} p(\lambda_k \mid X, l_{n(k)}),
\end{align*}
where $n(k)$ is the data association function that tells the index of the reference landmark associated with the $k$-th online landmark.
To evaluate the above equation, we need to define a measurement model
\begin{align*}
p(\lambda_k \mid X, l_{n(k)}) \coloneqq \mathcal{N}(\norm{X \lambda_k - l_{n(k)}}, \sigma) + \epsilon,
\end{align*}
with the reference and online landmarks represented by homogeneous 2\nobreakdash-D position vectors~\mbox{$\transpose{[x, y, 1]}$}, and where we assume isotropic position uncertainty~$\sigma$ of the reference landmarks.
The constant addend~\mbox{$\epsilon \in \mathbb{R^+}$} accounts for the probability of discovering a pole that is not part of the map.
This probability can be estimated by generating a global map from one run, generating a set of local maps from data recorded on the same trajectory in a second run, and computing the numbers of matched and unmatched landmarks.

\section{Experiments}
\label{sec:experiments}

In order to evaluate the proposed localization system, we perform two series of experiments.
The complete implementation is publicly available~\cite{polex2019}.
In the first series, we assess the system's long-term localization reliability and accuracy on the NCLT dataset~\cite{carlevaris2015}.
While these experiments provide profound insights into the performance of the developed method, the results are absolute and do not allow direct comparisons with other methods, because to the best of our knowledge, we are the first to test landmark-based localization on NCLT.
For this reason, we base the second experiment series on the KITTI dataset~\cite{geiger2013}.
That allows us to repeat the experiments performed by the authors of another state-of-the-art localization method, only that this time, we use the system presented above.

\subsection{Localization on the NCLT Dataset}
\label{subsec:localization_nclt_dataset}

The NCLT (North Campus Long-Term) dataset~\cite{carlevaris2015} was acquired with a two-wheeled Segway robot on one of the campuses of the University of Michigan, USA.
The data is perfectly suited for testing the capabilities of any system that targets long-term localization in urban environments:
Equipped with a Velodyne HDL-32E lidar, GPS, IMU, wheel encoders, and a gyroscope, among others, the robot recorded 27~trajectories with an average length of \SI{5.5}{km} and an average duration of \SI{1.3}{h} over the course of 15~months.
The recordings include different times of day, different weather conditions, seasonal changes, indoor and outdoor environments, lots of dynamic objects like people and moving furniture, and two large construction projects that evolve constantly.
Although the routes differ significantly between sessions, the trajectories have a large overlap.

The main difference between NCLT and the data used to evaluate all other pole-based localization methods we surveyed lies in its extent:
While related works briefly demonstrate the plausibility of their approaches by evaluating localization performance on datasets with durations between \SI{46}{s} and \SI{30}{min}, we focus on long-term reliability and accuracy and process \SI{35}{h} of data spread over more than one year.

\begin{figure*}
	\centering
	\begin{subfigure}[t]{0.49\linewidth}
		\includegraphics[width=\linewidth]{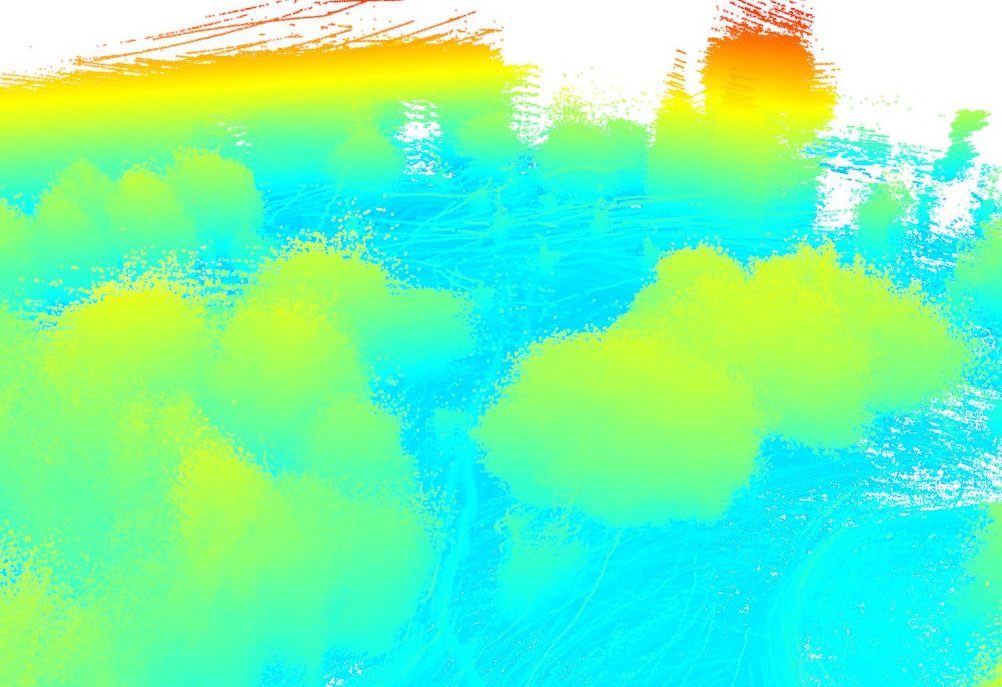}
		\caption{Registration via the original NCLT ground truth.}
		\label{fig:synpeb_seg_gt}
	\end{subfigure}
	\hfill
	\begin{subfigure}[t]{0.49\linewidth}
		\includegraphics[width=\linewidth]{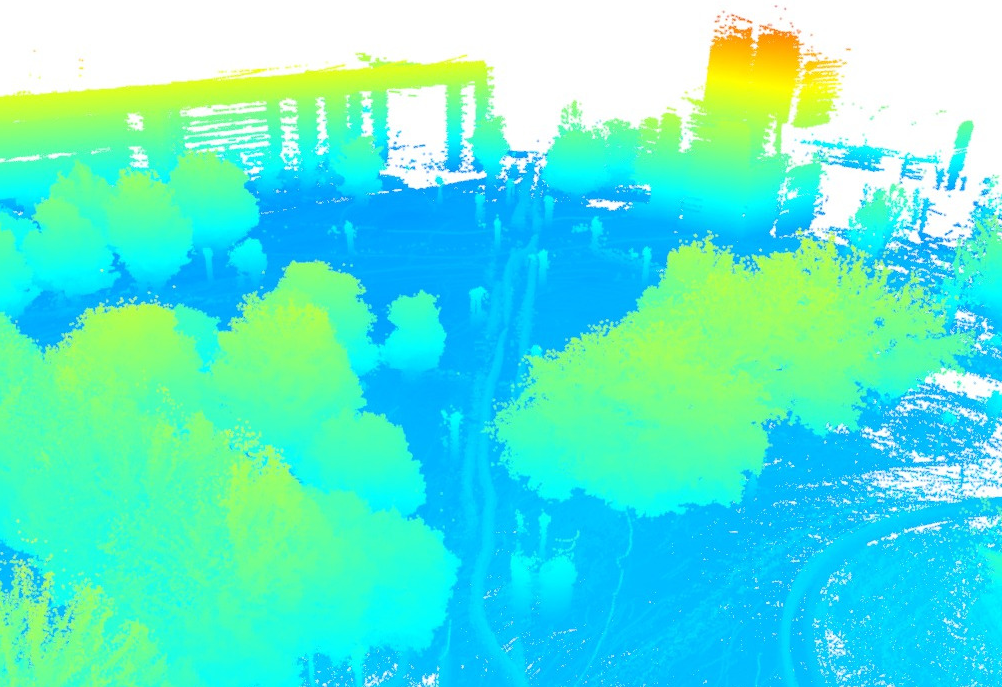}
		\caption{Refined registration.}
		\label{fig:synpeb_seg_ppe}
	\end{subfigure} 
	\caption{
		The same set of point clouds taken from a short sequence of an NCLT session, registered using different ground-truth robot poses.
		The colors encode the point height above ground: 
		Blue represents the ground plane, whereas green, yellow, and red indicate increasing height.
		The left image shows the result of the registration based on the original NCLT ground truth poses, which we use throughout our experiments.
		To illustrate the inaccuracy of the original ground truth, the right-hand side image presents a refined registration that we generated via pose-graph optimization.
		While the original ground truth leads to a blurry point cloud, the refined version significantly improves point alignment and results in crisp details.
		The mean positional error between both ground truth versions is approximately \SI{0.25}{m} on average, which leads us to believe that the original NCLT ground truth is off by a similar amount.
		This fact impedes the generation of an accurate reference pole map and negatively affects our localization results.}
	\label{fig:nclt_noise}
\end{figure*}

\begin{figure}
	\centering
	\includegraphics[width=\linewidth]{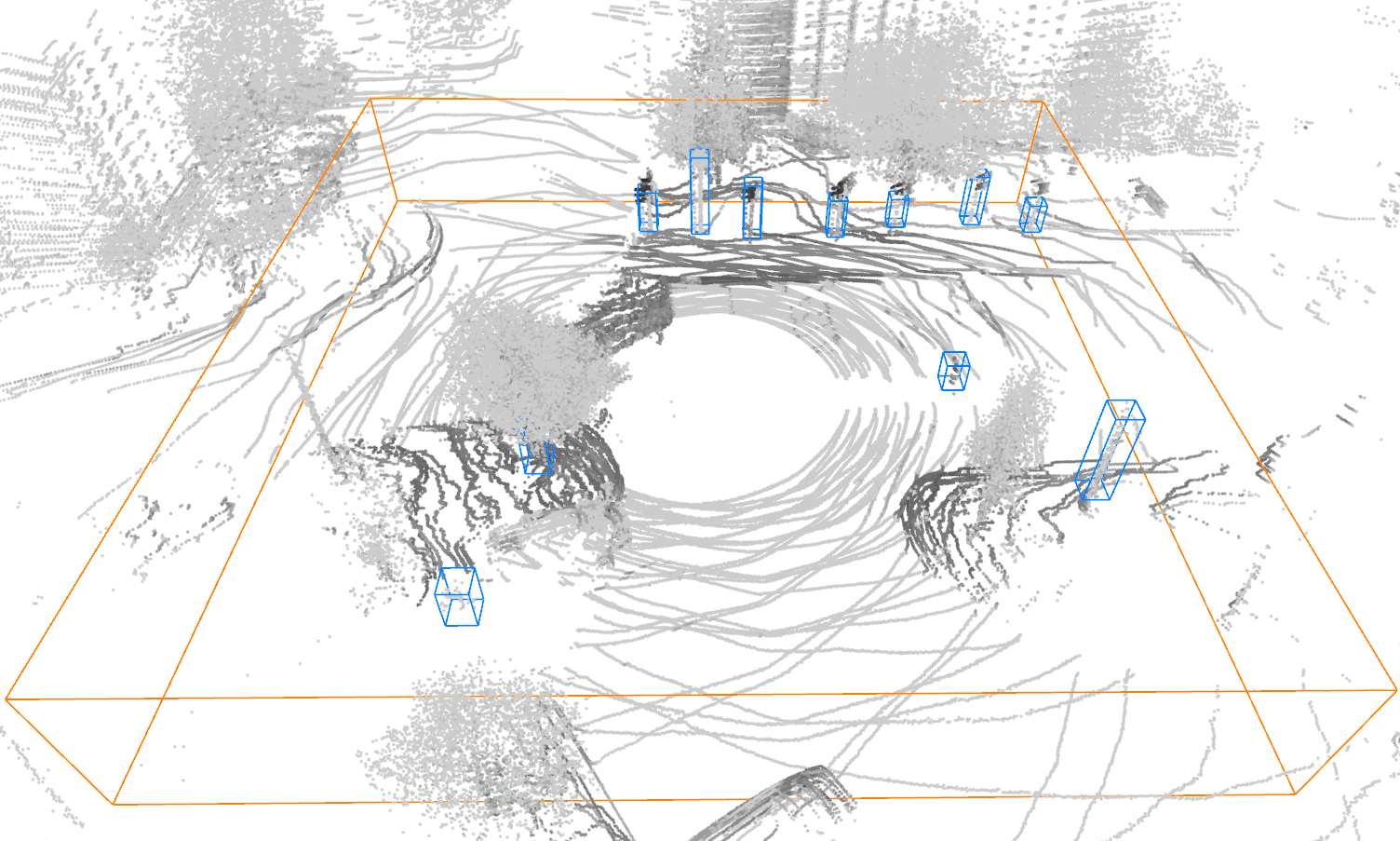}
	\caption{
		Exemplary pole extraction result for a point cloud from the NCLT dataset.
		The gray values of the points correlate with the intensity values returned by the lidar sensor.
		The orange wireframe represents the boundaries of the local map, while the blue wireframes represent the extracted poles.
		The pole extractor is triggered by different kinds of pole-shaped objects like traffic signs, street lamps, and tree trunks.}
	\label{fig:nclt_poles}
\end{figure}

Before localizing, we build a reference map of the poles on the campus.
To that end, we feed the laser scans and the ground-truth robot poses of the very first session to our mapping module.
Unfortunately, the ground truth provided by NCLT is not perfect.
It consists of optimized poses spaced in intervals of \SI{8}{m}, interpolated by odometry.
Consequently, point clouds accumulated over a few meters exhibit considerable noise, as illustrated in figure~\ref{fig:nclt_noise}.
For that reason, we set the distance of the trajectory segments to build local maps to \SI{1.5}{m}, the raster spacing of the grid maps to \SI{0.2}{m}, and the occupancy threshold to \mbox{$\mu_{o} = 0.2$}.
During mapping and localization, the pole extractor discards all poles below a minimum pole height of $h_{\min} = \SI{1}{m}$ and below a minimum pole score of $q_{\min} = 0.6$.
The extent of the local maps is chosen $\SI{30}{m} \times \SI{30}{m} \times \SI{5}{m}$ in $x$, $y$, and $z$ of the map frame, respectively.
Figure~\ref{fig:nclt_poles} illustrates the corresponding results. 

Although the first session covers most of the campus, the robot occasionally roams into unseen regions during later sessions.
For that reason, we iterate over all subsequent sessions, too, but add landmarks to the global map only if the corresponding laser scans are recorded at a minimum distance of \SI{10}{m} from all previously visited poses.
Table~\ref{tab:nclt_results} shows that after the second session, the fractions of scans per session that contribute to the map drop to $f_{\textrm{map}} \leq \SI{5.5}{\percent}$.

\begin{table}
	\centering
	\small
	\setlength{\tabcolsep}{3pt}
	\begin{tabular}{l
			S[round-mode=places, round-precision=1]
			S[round-mode=places, round-precision=3]
			S[round-mode=places, round-precision=3]
			S[round-mode=places, round-precision=3]
			S[round-mode=places, round-precision=3]}
		\toprule
		Session Date 
			& $f_{\textrm{map}}$
			& $\Delta_{\textrm{pos}}$
			& $\RMSE_{\textrm{pos}}$
			& $\Delta_{\textrm{ang}}$
			& $\RMSE_{\textrm{ang}}$ \\
			& [\si{\percent}]
			& [\si{m}]
			& [\si{m}]
			& [\si{\degree}]
			& [\si{\degree}] \\
		\midrule
		2012-01-08 & 100.0 & 0.130018287924 & 0.178063588239 & 0.662604678187 & 0.856942505413 \\
		2012-01-15 & 8.46853977048 & 0.155940689894 & 0.225168620856 & 0.759810914295 & 0.999075263741 \\
		2012-01-22 & 5.09090909091 & 0.171668372127 & 0.222314294622 & 0.938812315825 & 1.29130146491 \\
		2012-02-02 & 0.355956336023 & 0.155187456675 & 0.204719053305 & 0.720106471995 & 0.974942624438 \\
		2012-02-04 & 0.106298166357 & 0.14448874358 & 0.195382609041 & 0.683799096887 & 0.903184232846 \\
		2012-02-05 & 0.54114994363 & 0.14782774588 & 0.210140923771 & 0.690863940378 & 0.946520431301 \\
		2012-02-12 & 0.822833633325 & 0.268658751254 & 1.00466609751 & 0.802008569954 & 1.03989041028 \\
		2012-02-18 & 0.815738963532 & 0.148798848112 & 0.2212811784 & 0.699380463682 & 0.9378920653 \\
		2012-02-19 & 0.048111618956 & 0.148364622352 & 0.193940274986 & 0.70399833233 & 0.943677840762 \\
		2012-03-17 & 0.0 & 0.148822138842 & 0.190709677635 & 0.830459766755 & 1.06226575502 \\
		2012-03-25 & 0.0 & 0.20040549833 & 0.261677204157 & 1.41775364206 & 1.83560766223 \\
		2012-03-31 & 0.0 & 0.142955070466 & 0.184486716013 & 0.745526184794 & 0.97312573778 \\
		2012-04-29 & 0.0 & 0.169516441578 & 0.251473668463 & 0.828732734741 & 1.07884729623 \\
		2012-05-11 & 5.46702623192 & 0.16147768668 & 0.225155811193 & 0.772501499396 & 0.997873058189 \\
		2012-05-26 & 0.378340033105 & 0.158296999211 & 0.217380143114 & 0.690064146222 & 0.88903376467 \\
		2012-06-15 & 0.403669724771 & 0.180027133802 & 0.23768580373 & 0.658830818121 & 0.879060612331 \\
		2012-08-04 & 0.273000273 & 0.209734982198 & 0.339523277926 & 0.88447514082 & 1.14268314933 \\
		2012-08-20 & 3.81451009723 & 0.18915365794 & 0.263847014715 & 0.710947440445 & 0.941364638647 \\
		2012-09-28 & 0.34965034965 & 0.206240646232 & 0.310817013818 & 0.730937439078 & 0.951629755132 \\
		2012-10-28 & 1.42405063291 & 0.217087955819 & 0.338443163605 & 0.69298500232 & 0.918813494296 \\
		2012-11-04 & 2.53441802253 & 0.25650337883 & 0.456421468716 & 0.745505789407 & 0.995694822974 \\
		2012-11-16 & 2.72922416437 & 0.4033767283 & 0.722437763064 & 1.46684092555 & 2.03114919862 \\
		2012-11-17 & 0.364678301641 & 0.242981573807 & 0.377457015298 & 0.685714080867 & 0.959180156256 \\
		2012-12-01 & 0.0 & 0.266108523719 & 0.492351132772 & 0.674320870805 & 0.930365440703 \\
		2013-01-10 & 0.0 & 0.217325612752 & 0.278399109185 & 0.689157726687 & 0.910747819784 \\
		2013-02-23 & 0.0 & 2.46959449558 & 5.47978388501 & 1.08266736608 & 1.76939797944 \\
		2013-04-05 & 0.0 & 0.365161620261 & 0.92031777341 & 0.654361646184 & 1.02792023674 \\
		\bottomrule
	\end{tabular}
	\caption{
		Results of our experiments with the NCLT dataset, averaged over ten localization runs per session.
		The variables $\Delta_{\textrm{pos}}$ and $\Delta_{\textrm{ang}}$ denote the mean absolute errors in position and heading, respectively, $\RMSE_{\textrm{pos}}$ and $\RMSE_{\textrm{ang}}$ represent the corresponding root mean squared errors, while $f_{\textrm{map}}$ denotes the fraction of lidar scans per session used to build the  reference map.}
	\label{tab:nclt_results}
\end{table}

During localization, odometry mean and covariance estimates are generated by fusing wheel encoder readings, gyroscope, and IMU data in an extended Kalman filter.
The particle filter contains 5000~particles, which we initialize by uniformly sampling positions in a circle with radius~\SI{2.5}{m} around the earliest ground-truth pose.
The headings are uniformly sampled in \mbox{$[\SI{-5}{\degree}, \SI{5}{\degree}]$}.
To maximize reliability, we inflate the motion noise by a factor of four, which corresponds to doubled standard deviation, define the position uncertainty of the poles in the global map as \mbox{$\sigma = \SI{1}{m^2}$}, and set the addend in the measurement probability density to \mbox{$\epsilon = 0.1$}.
We resample particles whenever the number of effective particles \mbox{$n_{\textrm{eff}} \coloneqq (\sum_{i} w_i^2)^{-1} < 0.5$}, where $w_i$ is the weight of the $i$-th particle, via low-variance resampling as described by Thrun et~al.~\cite{thrun2005}.
In order to obtain the pose estimate, we select the best \SI{10}{\percent} of the particles and compute the weighted average of their poses.

Table~\ref{tab:nclt_results} presents for each of the 27~sessions the corresponding position and heading errors.
To generate these values, we run the localization module ten times per session, evaluate the deviation of our estimate from ground truth every \SI{1}{m} along the ground-truth trajectory, compute the means and RMSEs, and average these metrics over the ten sessions.
The results demonstrate that the proposed method achieves both high reliability and accuracy, even if the data used for mapping and for localization lie 15~months apart:
The particle filter never even partially diverges, except for one late session discussed below.
Furthermore, despite the inaccuracies in ground truth, which affect both the global map and the evaluation, it achieves a mean positioning accuracy over all sessions of \SI{0.284}{m}.
Looking at the evolution of the errors over time, we observe slightly increasing magnitudes.
This is due to changes in campus infrastructure accumulating over time and rendering the initial map more and more outdated.

In session 2012-02-23, these changes eventually cause the localization module to temporarily lose track of the exact robot position.
The diverging behavior reproducibly occurs when the robot drives along a row of construction barrels that fence a large construction site.
When the global map was built, these barrels were located on the footpath.
Just before the session in question, however, the barrels were moved laterally by a few meters, while maintaining their longitudinal positions.
Since the barrels are the only landmarks in the corresponding region, the localizer ``corrects'' the robot position so that the incoming pole measurements match the map.
Having passed the construction site, the localizer is confident about its wrong position estimate, which is why it takes some time until the particle cloud  diverges and the robot relocalizes.
The positioning error over all sessions except 2012-02-23 amounts to \SI{0.200}{m}.

Lastly, we describe the runtime requirements of our method stochastically.
On a 2011 quad-core PC with dedicated GPU, we measure an average \SI{1.33}{s} for pole extraction with our open-source Python implementation~\cite{polex2019}, which corresponds to processing 0.5~million laser data points per seconds.
The measurement step with data association requires a mean computation time of \SI{0.09}{s}.
These two steps pose by far the highest computational requirements and make others, like the measurement update, negligible.

\subsection{Localization on the KITTI Dataset}
\label{subsec:localization_kitti_dataset}

As delineated in section~\ref{sec:related_work}, K{\"u}mmerle et~al.~\cite{kuemmerle2019}, Weng et~al.~\cite{weng2018}, and Sefati et~al.~\cite{sefati2017} present methods for vehicle localization with pole landmarks extracted from 3\nobreakdash-D lidar data.
While the former two use small proprietary datasets -- a fact that makes a direct comparison infeasible -- Sefati et~al. evaluate their method on sequence number~9 of the publicly available KITTI dataset~\cite{geiger2013}.
This sequence is a short recording of \SI{46}{s} along a simple L-shaped trajectory.
Trajectories in KITTI have hardly any overlap, which is why Sefati et~al. use the sequence data for both mapping and localization.
Consequently, their results have limited significance as to real localization performance:
They could theoretically localize the vehicle based on dynamic landmarks only, and they would still obtain accurate results with respect to their map, although it is extremely unlikely that they will encounter the same constellation of dynamic objects ever again.
The same is true for Weng et~al., who also use a single trajectory of \SI{3.5}{km} for mapping and localization.
Nevertheless, we repeat Sefati et al.'s experiment with the localization system proposed in this paper and compare accuracies in table~\ref{tab:kitti_results}.
This time, we set the grid spacing for the pole extractor to \SI{0.1}{m}, because the quality of the ground-truth robot poses is higher than in NCLT.
Furthermore, we adjust the parameters of our localizer to match the values Sefati et~al. apparently used -- 2000 particles, \SI{3}{m} initial positioning variation, \SI{\pm 5}{\degree} heading variation -- and average our results over 50~experimental runs.
As shown in table~\ref{tab:kitti_results}, our localization system outperforms the reference method by reducing the RMSEs in position and heading by \SI{54}{\percent} and \SI{69}{\percent}, respectively.
For qualitative analysis, table~\ref{tab:kitti_results} also includes the results K{\"u}mmerle et~al. and Weng et~al. obtained after processing their respective proprietary datasets.

\begin{table*}
	\centering
	\begin{tabular}{lccccccccc}
		\toprule
		Approach
			& $\Delta_{\textrm{pos}}$
			& $\RMSE_{\textrm{pos}}$
			& $\Delta_{\textrm{lat}}$
			& $\sigma_{\textrm{lat}}$
			& $\Delta_{\textrm{lon}}$
			& $\sigma_{\textrm{lon}}$
			& $\Delta_{\textrm{ang}}$
			& $\sigma_{\textrm{ang}}$
			& $\RMSE_{\textrm{ang}}$ \\
			& [\si{m}]
			& [\si{m}]
			& [\si{m}]
			& [\si{m}]
			& [\si{m}]
			& [\si{m}]
			& [\si{\degree}]
			& [\si{\degree}]
			& [\si{\degree}] \\
		\midrule
		K{\"u}mmerle et~al.~\cite{kuemmerle2019}
			& \tablenum[table-format=1.3]{0.12}
			& ---
			& \tablenum[table-format=1.3]{0.07}
			& ---
			& \tablenum[table-format=1.3]{0.08}
			& ---
			& \tablenum[table-format=1.3]{0.33}
			& ---
			& --- \\
		Weng et~al.~\cite{weng2018}
			& ---
			& ---
			& ---
			& \tablenum[table-format=1.3]{0.082}
			& ---
			& \tablenum[table-format=1.3]{0.164}
			& ---
			& \tablenum[table-format=1.3]{0.329}
			& --- \\
		Sefati et~al.~\cite{sefati2017}
			& ---
			& \tablenum[table-format=1.3]{0.24}
			& ---
			& ---
			& ---
			& ---
			& ---
			& ---
			& \tablenum[table-format=1.3]{0.68} \\
		Ours
			& \tablenum[table-format=1.3]{0.096}
			& \tablenum[table-format=1.3]{0.111}
			& \tablenum[table-format=1.3]{0.061}
			& \tablenum[table-format=1.3]{0.075}
			& \tablenum[table-format=1.3]{0.060}
			& \tablenum[table-format=1.3]{0.067}
			& \tablenum[table-format=1.3]{0.133}
			& \tablenum[table-format=1.3]{0.188}
			& \tablenum[table-format=1.3]{0.214} \\
		\bottomrule
	\end{tabular}
	\caption{
		Comparison of the accuracies of Sefati et~al.'s method and the proposed localization approach on the KITTI dataset.
		The results of Weng et~al. and K{\"u}mmerle et~al. are not directly comparable and are stated for qualitative analysis only.}
	\label{tab:kitti_results}
\end{table*}

\section{Conclusion and Future Work}
\label{sec:conclusion}

We presented a complete landmark-based 2\nobreakdash-D localization system that relies on poles extracted from 3\nobreakdash-D lidar data, that is able to perform long-term localization reliably, and that outperforms current state-of-the-art approaches in terms of accuracy.
The implementation is publicly available~\cite{polex2019}.

For the future, we have two major extensions in mind.
First, we plan to fuse the separated mapping and localization modules into a single SLAM module.
Second, we would like to explore pole-based localization in different sensor modalities.

\section{Acknowledgements}

We thank Arash Ushani for his kind support with the NCLT dataset.

\bibliographystyle{IEEEtran}
\bibliography{IEEEabrv,\jobname}

\end{document}